\documentclass{article} % For LaTeX2e
\usepackage{colm2024_conference}

\usepackage{microtype}
\usepackage{hyperref}
\usepackage{url}
\usepackage{subfigure}
\usepackage{booktabs}
\usepackage{todonotes}
\usepackage{amsmath}
\usepackage{amssymb}
\usepackage{mathtools}
\usepackage{amsthm}
\usepackage{adjustbox}
\usepackage{caption}
\usepackage{subcaption}
\usepackage{xspace}
\usepackage{float}
\usepackage{algorithm,algorithmic}
\usepackage{wrapfig}
\usepackage{diagbox}

\title{Toward Inference-optimal Mixture-of-Expert  Large Language Models}

\author{Longfei Yun$^{\text{1}}$\footnotemark[1]\\
\And
Yonghao Zhuang$^{\text{2}}$\footnotemark[1] \\
\And
Yao Fu$^{\text{3}}$ \\
\And
Eric P Xing$^{\text{2}}$ \\
\And 
Hao Zhang$^{\text{1}}$ \\
}

\begin{document}
\maketitle
\renewcommand{\thefootnote}{\fnsymbol{footnote}}
\footnotetext[1]{Equal contribution. \enskip $^{\text{1}}$UC San Diego \enskip $^{\text{2}}$Carnegie Mellon University \enskip $^{\text{3}}$The University of Edinburgh}

\maketitle

\begin{abstract}
% Several recent advances in large language models are a direct result of scaling the model. To continue scaling model size without suffering from the corresponding quadratic growth of training cost, 
Mixture-of-Expert (MoE) based large language models (LLMs), such as the recent Mixtral and DeepSeek-MoE, have shown great promise in scaling model size without suffering from the quadratic growth of training cost of dense transformers. 
% is getting more attention. MoE scales model size by adding ``experts'' operating in parallel. It avoids additional computation by only activating a subset of experts for a token.
Like dense models, training MoEs requires answering the same question: given a training budget, what is the optimal allocation on the model size and number of tokens? We study the scaling law of MoE-based LLMs regarding the relations between the model performance, model size, dataset size, and the expert degree. Echoing previous research studying MoE in different contexts, we observe the diminishing return of increasing the number of experts, but this seems to suggest we should scale the number of experts until saturation, as the training cost would remain constant, which is problematic during inference time. We propose to amend the scaling law of MoE by introducing inference efficiency as another metric besides the validation loss. We find that MoEs with a few (4/8) experts are the most serving efficient solution under the same performance, but costs 2.5-3.5x more in training. On the other hand, training a (16/32) expert MoE much smaller (70-85\%) than the loss-optimal solution, but with a larger training dataset is a promising setup under a training budget.
\end{abstract}

\section{Introduction}
% Recent practice have shown Mixture-of-Expert (MoE) models can significantly outperform Dense Transformers. MoE consists a group of experts. Each token in the input is only routed to a fixed size subset of all experts, and their outputs are integrated. When increasing the number of experts, the model's computation cost remains the same, because the number of activated parameters for each token is fixed. Such a sparsity leads to a `free lunch': with the same pre-training cost, more experts bring the model a potential to store more knowledge in the extra parameters. However, does scaling MoE on the number of experts as good as it sounds? In this paper, we study the optimal configuration of the number of experts for MoE models by considering two aspects: the scaling behavior for MoE, and the inference efficiency.

% 
Recent developments, such as Mixtral~\citep{jiang2024mixtral}, DeepSeek-MoE~\citep{dai2024deepseekmoe}, spotlight Mixture-of-Experts (MoE) models as a superior alternative to Dense Transformers. 
% An MoE layer comprises a group of experts, where each input token is routed to a fixed number of these experts, and their outputs are subsequently integrated. 
An MoE layer works by routing each input token to a selected group of experts for processing. Remarkably, increasing the number of experts in an MoE model (almost) does not raise the computational cost, enabling the model to incorporate more knowledge through extra parameters without inflating pre-training expenses. This approach seemingly presents a ``free lunch'' that we could just infinitely scale the number of experts -- yet raises a critical question: is scaling up the number of experts in MoE models always as beneficial as it seems? In this paper, we answer this question and investigate the optimal number of experts for MoEs by examining two key factors: the scaling behavior and inference efficiency.

% To understand how scaling an MoE model improves its performance, we first study the scaling behavior for MoE model. 
To understand how performance improves when scaling MoE models, we first study its scaling behavior. Previous works on Transformer model~\citep{kaplan2020scaling,hoffmann2022training} have established a power-law relationship linking the model’s validation loss $L$ to both the number of parameters $N$ and training tokens $D$, which is referred to as the scaling law. Together with an estimation of training cost $C(N,D)$, there is an optimal $(N,D)$ within a training budget $C_0$ (i.e., $\arg\min L(N,D),s.t.\ C(N,D)\le C_0$). We name this configuration a \emph{loss-optimal budget allocation}.

Our \emph{first contribution} is to enhance the existing scaling law to incorporate the number of experts $E$. Existing works either do not study the scaling behavior against $E$, or ignore the influence of the number of training tokens $D$ -- both are crucial to optimize the training budget allocation for MoEs. Akin to the Transformer scaling law, we observe that the number of expert, model size, and training dataset size all conform to a power-law relationship with validation loss. Consistent with previous work~\citep{clark2022unified}, our MoE scaling law also reveals a diminishing return for increasing the number of experts, which saturates at a threshold $E_{\max}$.

% Though the saturation of number of expert $E$ seems to answer the training budget allocation question, it is not a practical solution.
% The scaling law of MoE suggests a configuration with $E_{\max}$ experts, which leads to the optimal loss under a training budget.
Although our findings suggest a loss-optimal configuration with $E_{\max}$ experts, such a setup is not practical for actual deployment. The main reason is that an excessive number of experts makes the model impractical for inference. In contrast to pretraining, LLM inference is notably memory-intensive, as it requires storing intermediate states (KV-cache) of all tokens. With more experts, the available memory for storing KV caches is squeezed. As a result, the batch size -- hence throughput -- decreases, leading to increased cost per query. This observation suggests scaling MoE must be subject to inference cost.
% maybe a figure here? Show that when doubling the expert, how cost per token and model performance changes. e.g. one is linear, while the other is sublinear.
Our \emph{second contribution} is to incorporate inference cost, characterized by a new metric -- cost per token -- as a novel constraint for budget allocation for MoE models, in addition to the validation loss in  existing works~\footnote{In dense models, we cannot scale the number of parameters without increasing the training cost, hence the inference cost is predetermined and need not be separately considered in its scaling law.}. This dual-metric approach allows for a more comprehensive evaluation balancing model quality with practical resource constraints. % maybe why we use cost per token, compare with previous works on total inference dollars estimation

By jointly considering the scaling behavior under inference efficiency constraints, we first study loss-optimal models with different numbers of experts. We found that MoE models with 4 or 8 experts exhibit more efficient inference and higher performance compared to MoE models with more experts. However, they necessitate 2.4x-4.3x more training budgets to reach the same performance with models with more experts, making them impractical from the training side.

We further notice that for MoE with more experts, given a training budget, when the model shrinks a lot from the loss-optimal size, the performance only experience a marginal change. On the other hand, the inference cost grows linearly with the model size, and benefits a lot from a smaller model. This observation motivates us to train a model much smaller than the loss-optimal configuration. Such a model, though suffers from a marginal drop in quality, has a significantly lower inference cost. Because the budget saved from using a smaller model can be utilized to train on more tokens, we refer to this as an over-trained configuration. To evaluate the potential of over-trained models with more experts, we compare them with loss-optimal models with fewer experts under the same training budget. Under the same quality of a loss-optimal 4-expert MoE, an over-trained 8- or 16-expert MoE only needs 47.0\% to 52.0\% inference cost. With the same inference cost, an over-trained 16-expert MoE can save up to 68.4\% training budget.

Our main contributions can be summarized as follows:
\begin{itemize}
    \item We study the scaling law of MoE LLMs, revealing the relation between the validation loss and all 3 critical factors: model size, dataset size, and number of experts;
    \item We introduce a novel perspective to analyze the optimal training budget allocation for MoE models, which considers inference cost as a key component;
    \item We demonstrate that a smaller, over-trained MoE model with additional experts can surpass larger, fewer expert models in both quality and inference efficiency.
    
\end{itemize}

\section{Background}

\subsection{Mixture of Expert Model}
\label{sec:mixture-of-expert-model}
% Although scaling model size brings nontrivial improvement to Transformer models, it correspondingly introduces more computation, which bottlenecks the scaling ability. 
Many works on sparse models~\citep{jacobs1991adaptive, jordan1994hierarchical, shazeer2017outrageously, lepikhin2020gshard, fedus2022switch} have been introduced to continue scaling the sizes of large language models with a marginal increase on compute, among which, Mixture-of-Expert (MoE) is perhaps the most succesful example.
% MoE model is a variant of the transformer model, but introduces a sparsely-activate pattern to reduce the model's computation cost. 
% \todo{some of the above is duplicated from the intro.}
An MoE layer consists of a router and a set of experts. Every input token is routed to a subset of $K$ experts, and the outputs of these experts are combined to produce the final output (\autoref{fig:moe-arch}). 
It is common to replace the Feed-forward layer (FFN) in a Transformer model with MoE layers. The architecture of each expert is identical to the replaced FFN. 
%which is a two-layer MLP.
% Hence, the computation of an MoE layer is expressed as:
% \begin{equation*}
%     \begin{aligned} \mathbf{h}_t^l & =\sum_{i=1}^E\left(g_{i, t} \operatorname{FFN}_i\left({u}_t^l\right)\right)+{u}_t^l, \\ g_{i, t} & = \begin{cases}s_{i, t}, & s_{i, t} \in \operatorname{Topk}\left(\left\{s_{j, t} \mid 1 \leqslant j \leqslant E\right\}, K\right), \\ 0, & \text { otherwise }\end{cases} \\ s_{i, t} & =\operatorname{Softmax}_i\left({u}_t^{{l}^T} {e}_i^l\right),\end{aligned}
% \end{equation*}
% here $u_t^l$ is the hidden state of the token after the $l$-th attention module, $h_t^l$ represents the output, $g_{i,t}$ denotes the gate value for the $i$-th expert, $s_{i, t}$ represents the affinity between token and the $i$-th expert, and $e_i^l$ is the centroid of the $i$-th expert in the $l$-th layer. It's important to note that Topk refers to the group that contains the $K$ highest affinity scores. These scores are calculated for the 
% $t$-th token against all $E$ experts.

% Recent works~\cite{du2022glam, st_moe, fedus2022switch, lepikhin2020gshard} suggest to replace one of every two FFN layers in a Transformer model by MoE, and use Top-2 gating~\cite{shazeer2017outrageously} as the routing mechanism. In this paper, we also inherit from such a context.
A critical factor in MoEs is the number of parameters activated to process a single token. 
% To depict the two number, 
We introduce a notion \emph{Corresponding Dense Model}, which refers to a dense Transformer model with an identical number of layers and hidden dimension size as the MoE model. If the Corresponding Dense Model of an MoE has a size of $N$, its total activated number of parameters for a token is roughly $(Ka + (1-a))N$. Here $a$ is the proportion of the size of MLP layers relative to the size of the dense model. Since all components of the model scale simultaneously, $a$ is a constant for a given architecture.

\subsection{Scaling Law}
\label{sec:background_scaling_law}
Recent research~\citep{kaplan2020scaling, hoffmann2022training, brown2020language} indicate that scaling the number of parameters in a dense Transformer model or the size of the training dataset yields a predictable outcome on the model's final perplexity. Such correlation typically follows a power law of the parameters (N) and training tokens (D):
\begin{equation}
    L(N,D) = L_0 + \frac{A}{N^\alpha} + \frac{B}{D^\beta}
    \label{eq:openai_scaling_law}
\end{equation}
where $L_0,A,B,\alpha,\beta$ are constants whose values depend solely on the model architecture and the training data corpus, i.e. the quality of the dataset. 

A common practice to determine the most effective allocation of the training budget is to utilize scaling laws:
\begin{equation}
    \underset{N, D}{\operatorname{argmin}} L(N, D) \text { s.t. } \operatorname{FLOPs}(N, D)=C
    \label{eq:optimal_allocation}
\end{equation}
We refer to this choice of $(N,D)$ under the constraint as the \emph{loss-optimal} configuration.

% These constants are integral in defining how model architecture and the training dataset influence the model's overall performance.

% Based on the scaling law, for a given training budget $C$, 
% there is a corresponding configuration of $N$ and $D$ to have the lowest loss. This configuration is solved by taking $C=6ND$~\cite{kaplan2020scaling} into the formulation of $L(N,D)$ above. Here 6 is a constant based on the Transformer architecture. 
% We name this choice of $(N,D)$ a `loss-optimal' configuration.

% In addition to the loss-optimal configuration under a certain budget, the scaling law also reveals how much does the dataset size grow when the model size is doubled. By applying the training cost estimation $C=6ND$ to \autoref{eq:openai_scaling_law} and differentiating with respect to $N$ and $D$, there is:
They also propose to calculate the loss-optimal configuration as follows (\autoref{appendix:opt_alloc}): 
\begin{equation}
\begin{aligned} 
    N_{\text{opt}}(C)&=G\left(\frac{C}{6}\right)^a, \quad D_{\text{opt}}(C)=G^{-1}\left(\frac{C}{6}\right)^b,  \\
    \text{where } G& =\left(\frac{\alpha A}{\beta B}\right)^{\frac{1}{\alpha+\beta}}, a=\frac{\beta}{\alpha+\beta}, b=\frac{\alpha}{\alpha+\beta}
    \label{eq:openai_scaling_law_opt}
\end{aligned}
\end{equation}
Because $\alpha \approx \beta$, it is concluded that $N$ and $D$ should be scaled proportionally in compute-optimal training.

% The relationship between $a$ and $b$ reveals the expansion speed between model size $N_{\text{opt}}$ and number of tokens to train $D_{\text{opt}}$. 

% When $a$ is significantly larger than $b$, the model size scales faster than the training dataset size, and vise versa.

% Several studies have explored variations and extensions of the standard scaling laws of transformer models, examining how these modifications impact performance and efficiency. 
% \cite{muennighoff2023scaling} models the impact of limited data availability, where a model is trained for multiple epochs using the same dataset. 
% \cite{frantar2023scaling} investigates the scaling laws of sparsely-connected models, focusing on models where each parameter tensor undergoes either an unstructured or an n:m sparsity. This approach to sparsity differs fundamentally from the strategy employed in MoE models, where sparsity is achieved by activating selected model branches.
\cite{clark2022unified} explores the scaling behavior of MoE models. They introduce a multiplicative factor to capture the interaction between $N$ and $E$. Furthermore, they incorporate a saturation threshold ($E_{\max}$) to account for the diminishing returns observed when the number of experts ($E$) becomes excessively large. 
However, this study does not consider the impact of the dataset size ($D$) on the model's performance. As a result, it fails to provide recommendations on the loss-optimal configuration for a given training budget. 

%Besides, it does not address the scaling behavior of the well-adapted Top-2 gating mechanism, and neither give adequate attention to the inference cost.

\subsection{LLM Inference}
At inference, LLMs generate tokens following an auto-regressive paradigm. At the first iteration (prompt stage), the model generates the hidden states for all prompt tokens. 
%It then uses these hidden states to predict the first token of the response. 
In subsequent iterations (decoding stage), the model generates the hidden state for the most recently generated token and uses the accumulated hidden states to predict the next token. These hidden states, known as KV cache, are retrained in memory for compute efficiency.
During the decoding phase, each iteration merely computes the hidden state of one token per request, resulting in low compute intensity on accelerators. To minimize the cost per query, we want to batch many requests to boost the serving throughput. Consequently, the size of the cumulative KV cache across all requests, even with optimizations like Multi-query attention (MQA), is very large and becomes significantly memory-bound. Hence, the available memory to store KV caches dictates the batch size -- hence throughput and cost per query.
% KV cache becomes a critical memory bottleneck for increasing the batch size. 
% For instance, in the Llama-65B model, while the model parameters require 140GB of memory, a single request with 1024 tokens can consume as much as 2.5GB of memory solely for the KV-cache. 
% Many works like Multi-query attention (MQA) ~\citep{shazeer2019fast,ainslie2023gqa} reduce memory demands of the KV-cache by reducing its size by a factor less than or equal to 8.
% During the decoding phase, each iteration merely computes the hidden state of one token per request, resulting in low compute intensity, yet it necessitates loading all model parameters and storing the KV-cache of all tokens for each request.

% This phenomenon primarily stems from the low computational intensity of the decoding stage. 
 % Consequently, even when batching hundreds of requests, the latency increase remains marginal, and a cost-effective batch size is in the order of hundreds.
%Besides, to align the trade-off between training and inference, they 
% background on moe:
% define what is a base-model;
% moe's advantage: scale the #param, but keep the FLOPs, thus the training cost does not increase a lot
% moe's disadvantage: memory pressure on inference(maybe add a subsection to introduce auto-regressive inference is memory bound, and put moe's disadvantage there?)

% background on scaling law:
% 1. the variables to be considered(the dense model's size, dataset size, number of experts);
% 2. the budget-based variant of a scaling law. Mainly by a reversed function
\newcommand{\base}{\textsc{s-base}\xspace}
\newcommand{\baseorig}{\textsc{base}\xspace}
\newcommand{\rlr}{\mbox{\textsc{rl-r}}\xspace}
\newcommand{\hash}{\textsc{hash}\xspace}

\section{Method: Scaling law of MoE model}
\label{sec:scaling-law} 
Though the scaling law for dense Transformer is already well developed, it still lacks exploration in the context of MoE models. In this section, we develop the MoE's scaling law from some previous exploration~\citep{clark2022unified}. 

\subsection{Experiment setup}
To study the scaling behavior, we train a sweep of models with a dense model size ranging from 100 million to 730 million parameters. The detail of each model's hyper-parameters is in \autoref{tab:architectures}. For every dense model, we trained with 4, 8, 16, and 32 experts, with a dense Transformer as the baseline. We construct the training dataset by uniformly sampling from SlimPajama~\citep{cerebras2023slimpajama}, with a size ranging from 2.5B to 20B. More training details can be found in \autoref{appendix:training_details}.

\begin{table}[H]
    \centering
    \caption{Model Configurations}
    \vskip -0.9em
    \label{tab:architectures}
    \begin{tabular}{ c | c c c c c}
     \toprule
      \textbf{Name}
      & $d_{\textit{model}}$
      & $n_\mathit{layers}$
      & $n_\mathit{heads}$
      & Actual \# Params (w/o embedding)
      \\ [0.5ex] 
      \midrule
      100M & 768 & 12 & 8 & $81,395,712$ \\ 
      200M & 896 & 14 & 8 & $184,64,768$ \\ 
      320M & 1024 & 16 & 12 & $289,406,976$ \\ 
      730M & 1536 & 16 & 16 & $679,477,248$ \\ 
      %1.4B & 2048 & 32 & 16 & $1,214,251,008$ \\ 
      \bottomrule
    \end{tabular}
    \vskip -1em
\end{table}

\subsection{Formulate the scaling law for MoE}
Observations from \autoref{fig:fitted_val_loss} show parallel lines for various dense model sizes, indicating a consistent slope across all sizes. This uniformity in slope is also apparent under different numbers of training tokens, with the lines differing primarily in their intercepts.
On top of that, provided other factors do not become limiting, increasing the number of experts leads to a proportional decrease in validation loss. This trend holds true regardless of the dense model size and the number of training tokens used.

Based on the sweep of experimental runs, we observe a similar finding to the existing work~\citep{clark2022unified}, that not all models across $N$ benefit equally from $E$, though $E$ roughly follows a power-law to a certain extent. As a result, we inherit the interaction term of $N$ and $E$ from the existing work.

However, the relation between $D$ and $E$ has not yet been explored. From \autoref{fig:fitted_val_loss}, we observe that the benefit between two distinct number of experts $E$ remains constant across different numbers of tokens $D$, indicating that an interaction term between $D$ and $E$ is unnecessary. It is also reasonable to conjecture that, when the router's error rate is roughly the same, a fixed number of tokens are dispatched to the correct expert to be learned, regardless of the number of experts. We also find the same saturating trend: as $E$ increases, the benefit decreases, which is evident from scaling $E$ from 16 to 32.

As a result, building upon the existing research~\citep{kaplan2020scaling, clark2022unified} and experimental heuristics, we introduce a new scaling law that extends their theories to the MoE architecture:
{
\small
\begin{equation}
\begin{aligned}
    & \log {L(N, D, E)} \triangleq \log ( {\frac{A}{N ^ \alpha} + \frac{B}{\hat{E}^\beta} + \frac{C}{D^\gamma} + F} ) + d \log N \log \hat{E} \\
    & \text { where } \frac{1}{\widehat{E}} \triangleq \frac{1}{E-1+\left(\frac{1}{E_{\text {start }}}-\frac{1}{E_{\max }}\right)^{-1}}+\frac{1}{E_{\max }}
    \label{eq:moe_scaling_law}
\end{aligned}
\end{equation}
}
The first term represents the ideal performance achievable in a hypothetical space. However, the routing mechanism constrains the actual performance, leading to the introduction of the second term.
$E_{start}$ and $E_{max}$ are two terms fitted to model the saturation, ensuring that the scaling behavior is bounded on both sides.
$\hat{E}$ signifies that for $E\gg E_{start}$ and $E \ll E_{max}$, performance varies near-linearly. The peak performance is equivalent to the performance obtained with $E_{max}$ experts without saturation.

\subsection{Fit Result}
\begin{figure*}[t]
\vspace{-1em}
  \centering \includegraphics[width=\textwidth]{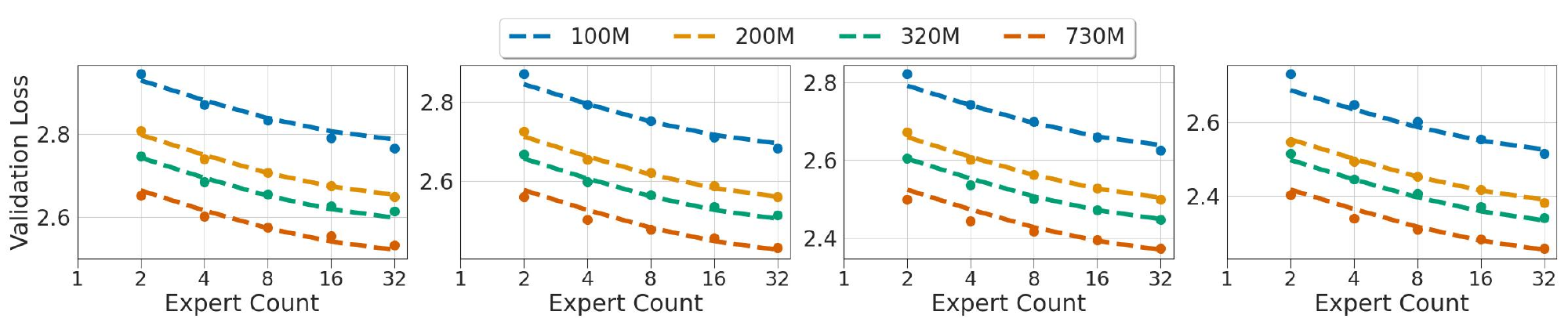}
      \hspace{3mm}(a) 5.0B Tokens \hspace{10mm} (b)  7.5B Tokens \hspace{10mm} (c) 10B Tokens \hspace{10mm} (d) 20B Tokens
\caption{\textbf{Validation losses for different $D$.} Scattered dots show the actual losses, and dotted lines correspond to values fitted by \autoref{eq:moe_scaling_law}.}
\label{fig:fitted_val_loss}
\vspace{-1em}
\end{figure*}
\autoref{fig:fitted_val_loss} displays the predicted outcomes derived from our scaling law. More details of the fitting procedure is in \autoref{sec:appendix-scaling-experiment-details}. 
The goodness-of-fit is evident, demonstrating that the fitted validation loss closely mirrors the actual validation loss. Therefore, \autoref{eq:moe_scaling_law} serves as an ideal model to represent the relationship between validation loss, model size, number of tokens, and number of experts. 

\section{Method: Estimating inference cost for MoE}
\label{sec:inference-cost}
% \begin{table}[htbp]
% 	\centering
% 	\footnotesize
%     \vskip -1em
% 	\caption{Notations.}
%     \vskip -1em
% 	\begin{tabular}{cc}
% 		\toprule
% 		Term & Symbol \\
% 		\midrule
%             hidden dimension and number of layers of model $m$ & $h_m, l_m$ \\
%             Parameter size of model $m$ & $N_m$ \\
%             Number of GPUs to serve the model & $G$ \\
%             Memory of a single GPU & $M_0$ \\
%             average prompt/answer length per request & $p$, $n$ \\
%             Maximal batch size of model $m$ & $b_m$ \\
%             Per iteration latency of model $m$ & $Lat_{m}$ \\
% 		\bottomrule
% 	\end{tabular}
%         \vskip -1em
% \end{table}

\label{sec:memory-bound}

Although scaling the number of experts in MoE models can procure higher performance without increasing the training budget, it incurs a significantly higher inference cost. Therefore, when determining the "optimal" number of experts, it is essential to consider the inference cost. In this section, we model the inference costs and analyze the MoE model's inference cost as the number of the experts increases.

\subsection{Inference cost estimation}

% As discussed in previous works\cite{narayanan2023cheaply}, the output generation time is linear to the number of output tokens. In other words, the latency to generate a single token is constant; 
As highlighted in prior research~\citep{narayanan2023cheaply}, there is a linear relationship between the time to generate output and the number of output tokens. In other words, the latency in generating each token remains consistent.
Thus, the throughput of a model $m$ is in the form of $T_m(N_m) = \frac{b_m(G)}{Lat_m(G, b)}$, 
where $G$ is the number of GPUs to serve the model, $b$ is the maximal batch size, and $Lat$ is the latency of a single iteration to generate a token. 
We derive the maximal batch size, latency and throughput in ~\autoref{appendix:inference_cost}.

% Not very related.
% Two techniques reduce such a memory pressure. The first is Group Query Attention(GQA). Originally, an attention head in the Key or Value tensor maps to a head in the Query tensor, while GQA makes it an 1-n mapping. This reduces the memory consumption for n times, only with a trivial performance drop. The other is PagedAttention. Instead of allocating the memory for all $2k$ tokens, it on demand allocates the memory of each request's hidden state. As most shown in the ShareGPT and LMSys trace~\cite{}, the average length of an answer is around 200 to 400, making the memory demand another 5 to 10 times smaller.

% However, even with the two optimizations, the average memory demand of a request is still at the scale of hundreds MB, making a 8x40GB A100 node only able to support about a thousand requests concurrently.

\paragraph{Inference cost} We define the inference cost in terms of dollars per token:
\begin{equation}
    C_{\text{Model},G} = \frac{GC_0}{T_{\text{Model}}(G)}
\label{eq:inference-cost}
\end{equation}
Here $C$ represents the cost per token, while $G$ denotes the number of GPUs utilized. $C_0$ is defined as the cost of operating a single GPU per second, which is usually considered a constant. Since the vendor has the flexibility to use any number of GPUs, we define $C(m)$ of model $m$ based on the most cost-effective GPU utilization, i.e., $C(m)=\min_{G}(C_{m,G})$, meaning we select the minimum cost across different GPU numbers for the most economical option of model $m$.

\subsection{MoE inference cost}
\label{sec:moe-inference-vs-dense}
As discussed in ~\autoref{appendix:inference_cost}, the size of an MoE model has $N_{MoE} = (1+(E-1)\cdot 1/3)N$. We take this term into \autoref{eq:inference-cost} and \autoref{eq:throughput}  to estimate the inference cost of MoE. 

% \begin{figure*}[h]
% \vspace{-2em}
% \centering     %%% not \center
% \subfigure[Inference cost under different dense model sizes.]{
% \includegraphics[width=0.4\textwidth]{images/plan9_v2.pdf}
%     \label{fig:inference-cost-with-model-size}
% }
% \subfigure[Maximal model size under a given inference cost.]{
% \includegraphics[width=0.4\textwidth]{images/model_max_size.png}
%     \label{fig:model_max_size}
% }
% \vspace{-1em}
% \caption{\textbf{MoE inference cost.} Cost increases proportionally with model size.}
% \vspace{-0.5em}
% \end{figure*}
\begin{figure*}[h]
\vspace{-2em}
\centering     %%% not \center
\includegraphics[width=.95\textwidth]{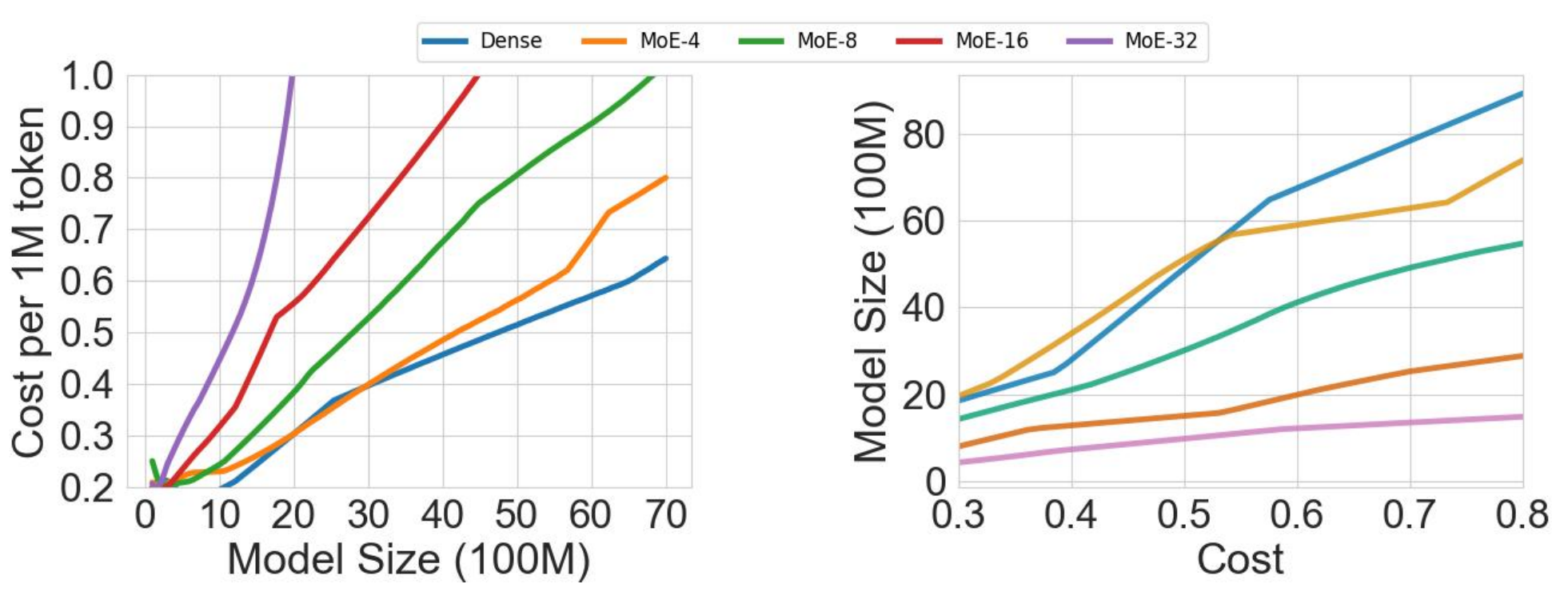}
\vspace{-1em}
\caption{\textbf{MoE inference cost.} Cost increases proportionally with model size.}
\label{fig:model_max_size}
\vspace{-0.5em}
\end{figure*}

In this paper, we profile the inference cost on 8x40 GB A100 GPU with NVLink connected, and use the state-of-the-art serving system vLLM~\citep{kwon2023efficient} to launch our model. 

~\autoref{fig:model_max_size} (left) shows the inference cost under different model sizes. Conversely, ~\autoref{fig:model_max_size} (right) plots the maximum model size for different inference budget. The relationship between inference cost and model size is mostly smooth and monotonic, and all exceptions occur because the minimum number of GPUs required to serve the model increases, which results in a gap in the inference cost.

\section{Results: Budget Allocation with Inference Efficiency}
\label{sec:scaling-behavior}
\begin{figure}[htbp]
\vspace{-0.5em}
  \centering \includegraphics[width=.95\columnwidth]{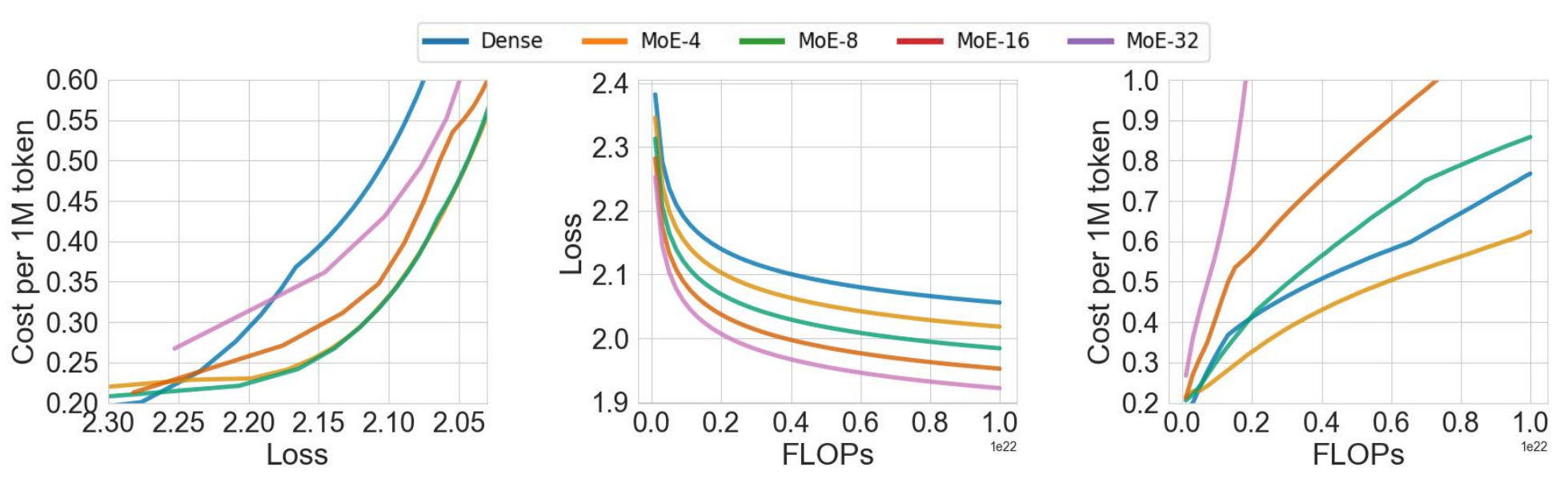}
\vspace{-1em}
\caption{\textbf{Trade-off between inference cost, model performance, and training cost.} Inference cost of and model performance for MoE models under different training budgets (left); Model performance with different training FLOPs (middle); Inference cost of different training FLOPs (right). Under the same budget, more experts means a better quality but higher inference cost. Fewer experts can reach a lower inference cost with the same quality, but needs much more training FLOPs}
\label{fig:training_inference_tradeoff}
\end{figure}

%%%%%%%%%%
%%%%%%%%%% Needs to tune
%%%%%%%%%%
Previous analysis already reveals a trade-off between inference cost and performance for MoE with different number of experts: on one hand, the scaling law (Section~\ref{sec:scaling-law}) shows that more experts (larger E) result in a higher performance; on the other hand, more experts result in a larger inference cost (Section~\ref{sec:moe-inference-vs-dense}). In this section, we first reveals another trade-off between training budget and inference cost (Section~\ref{sec:serving-centric-optimal-config}), then propose a budget allocation considering all these trade-offs (Section~\ref{sec:over-training}). The key idea of our purposed budget allocation is to relax the loss-optimal constraint during training, allowing a model with a sub-optimal performance, but a much lower inference cost.

 % Section~\ref{sec:moe-inference-vs-dense} shows that with a fixed training budget, 4-expert MoE outperforms dense Transformers on both inference cost and model performance. 
 % When comparing MoE model of different number of experts, there remains a trade-off between model performance and inference cost. When comparing the loss-optimal models under a fixed training budget, using more experts means a better performance but higher inference cost. In this section, we first relax the fixed training budget constraint (\ref{sec:serving-centric-optimal-config}), and find that a model perform better on inference cost needs much more training budgets. Then we relax the loss-optimal constraint, allowing the training budget not optimally utilized (\ref{sec:over-training}). Instead, a smaller model with more tokens is preferred, which results in a sub-optimal performance, but a much lower inference cost. Such \emph{over-training} configurations show the best trade-off between performance and cost.

\subsection{Trade-off between training and inference}
\label{sec:serving-centric-optimal-config}
For MoE model with different experts, there exists a trilemma among training budget, inference cost, and model quality. As shown in \autoref{fig:training_inference_tradeoff}(middle), for any fixed training budget, MoE with more experts have a higher performance (i.e., a lower loss). However, it suffers from a higher inference cost, as shown in \autoref{fig:training_inference_tradeoff}(right).

Since model training only runs for once, while model inference may serve unlimited number of requests, we also studies the correlation between the two inference metrics: model quality and inference cost. \autoref{fig:training_inference_tradeoff}(left) plots the model quality and inference cost under different training budgets, but guarantees that the model is loss-optimal. MoE with 4 or 8 experts shows the best quality (lowest validation loss) under a certain inference cost.

An explanation is that, the inference cost is approximately linear to the total number of parameters (\autoref{fig:model_max_size}). Under a fixed inference cost, if the number of experts is halved, the number of equivalent dense model's parameters is approximately doubled. For a loss-optimal configuration, the training dataset is scaled with the dense model's parameters, thus it is also doubled. In most cases, the gain of doubling both training dataset and the dense model's parameters outperforms the loss of halving the number of experts, and thus using fewer experts is more suggested.

However, since both the dense model and training dataset needs to be scaled up, MoE with fewer experts demands a much higher training budget to reach the same performance. By revisiting \autoref{fig:training_inference_tradeoff} (middle), there is a consistent trend that, to achieve the same loss with MoE models of fewer experts, it requires an increasing percentage in FLOPs. A 16-expert MoE only needs 23.7\% to 42.8\% of the FLOPs to reach the same model performance of a 4-expert MoE. When the total FLOPs increases, such a gap grows even larger.

This observation underscores that though MoE models with fewer experts (such as 4 or 8) consistently improve performance than more experts across both metrics on the inference side. However, this advantage comes at the cost of a much larger training cost.

\subsection{Over-training smaller MoE with more data}
% \begin{figure}[htbp!]
% \vspace{-1em}
%   \centering \includegraphics[scale=0.3]{images/plan2_v2.pdf}
% \vspace{-2.5em}
% \caption{inference cost and loss of loss-optimal models under different training budget.}
% \label{fig:loss-optimal-cost-with-flops}
% \end{figure}
\label{sec:over-training}
\begin{figure}[htbp!]
\vspace{-1.5em}
  \centering \includegraphics[scale=0.40]{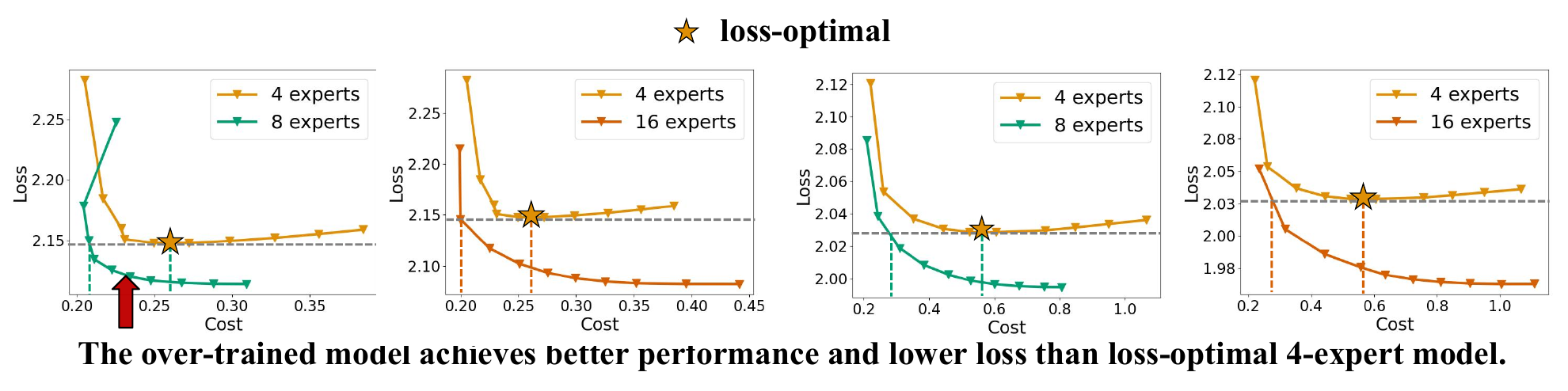}\\
\vspace{-0.5em}
  (a) 1e21FLOPs \hspace{40mm} (b) 8e21FLOPs
\vspace{-1.0em}
\caption{\textbf{loss-cost curve for a given training budget.} The over-trained 16-expert model achieves both better performance and lower inference cost than loss-optimal 4/8 expert model.}
\label{fig:cost_vs_loss_1e21}
\end{figure}

Though MoE of fewer experts has a lower inference cost, it needs an innegligible extra training budget. 
% However, as shown in \autoref{fig:cost_vs_loss_1e21}, given a training budget, the \emph{validation loss-inference cost} curve is very steep near the loss-optimal point. 
However, \autoref{fig:cost_vs_loss_1e21} demonstrates that, for a given training budget, the model with 8 or 16 experts outperforms the optimal 4-expert models in a specific region, achieving both improved performance and reduced cost.
This motivates us to consider such a case: what if we shift from the loss-optimal configuration to a model with fewer parameters, which leads to a much smaller inference cost? Since we can reuse the budget saved from model size to train more tokens, the model's quality only experiences a marginal drop within a range. We call this an \emph{over-trained} budget allocation. In this part, we study the potential of such over-trained model with more experts, and compare them with loss-optimal models with fewer experts under different scenarios.

More specifically, given a training budget $B$, we first find the loss-optimal budget allocation $(N_E,D_E)$ under a fixed number of experts $E$. The validation loss and inference cost for this model is correspondingly $L_E^{opt}$ and $I_E$. Then for MoE with a larger number of expert $E'>E$, we study its over-trained configuration, where its quality is anchored by $L_E$, say $L_{E'}\le L_E^{opt}$. We compute the lowest inference cost $I_{E'}^{\min}$ for $E'$ experts under the quality constraint above, and compare $I_{E'}^{\min}$ with $I_E$. On the other hand, we also consider the lowest validation loss $L_{E'}^{\min}$, under a bound that $I_{E'}\le I_E$, and compare $L_{E'}^{\min}$ with $L_E^{opt}$.

The practical meaning of the two is that, if an over-trained model reaches the quality of a loss-optimal model, can it have a lower inference cost? Or on the other hand, if the two model has the same inference cost, which one has a higher quality.

\paragraph{Optimal inference cost for a bounded loss.} 
Based on the scaling law, the loss $L$ is monotonic to model size $N$ before the loss-optimal point. Besides, the inference cost $I$ is also monotonic to $N$. As a result, to minimize inference cost $I$, the model size $N$ should be as low as possible, meaning the loss is as large as possible. As a result, $I_{E'}^{\min}$ corresponds to the case when the loss is exactly $L_E^{opt}$.

\begin{figure}[htbp]
\vspace{-1em}
  \centering \includegraphics[scale=0.41]{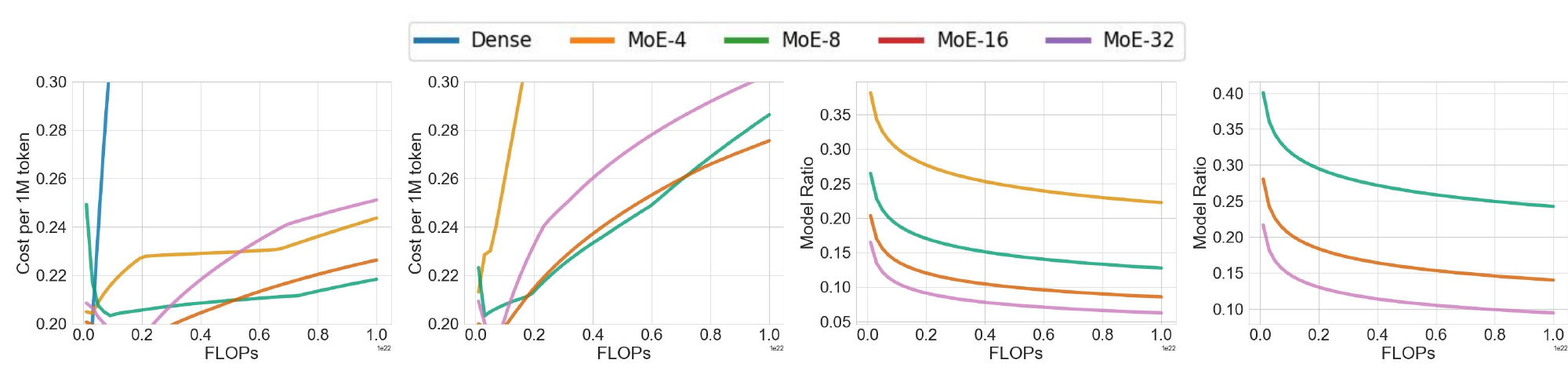} \\
  \vspace{-0.5em}
  (a) Base: Dense Model \hspace{6mm} (b) Base: MoE-4 \hspace{6mm} (c) Base: Dense Model \hspace{5mm} (d) Base: MoE-4
  \vspace{-0.5em}
\caption{\textbf{Optimal inference cost for a bounded loss.} Minimum achievable inference cost with a bounded loss (left). Ratio of model size to the base model (right).}
\label{fig:bound_loss}
\end{figure}

Based on the above analysis, we do dichotomy search for equation $L_{E'}(N,B)=L_E^{opt}$ to find the solution $N_{E'}$, and use it to compute $I_{E'}^{\min}$ (the detail is in Algorithm \ref{alg:bound_loss}). \autoref{fig:bound_loss} (left) shows the result for $E=1$ (dense Transformer) and $4$ (4-expert MoE). To reach the model performance (validation loss) similar to that of the dense model, over-training an 8-expert MoE with the same training budget has the lowest inference cost, which is 31.6\%-38.1\% as large as that of the dense model when $B$ ranges from 5.15e21 to 8.18e21.
When using 4-expert loss-optimal MoE's quality as a standard, 8-expert over-trained MoE saves 49.0\%-52.3\% inference cost per token, and 16-expert over-trained MoE saves 48\%-53\% inference cost. MoE with more experts has a higher cost than 8- or 16-expert. We reason this as that 4-expert MoE's optimal loss is too far away from 32-or-more expert MoE's, making the over-training no longer appealing as it already leaves the ``flat area'' in the size-loss curve.
 
% base4   ep8    ep16
% 5.15e21 0.510  0.520
% 8.18e21 0.477  0.470

% base      1    
% base_N    3.36B 
% ep8_ratio 16.9%     
% overtrained ep8 567M (2.12e21), optimal 2260M
% overtrained ep8 862M (5.95e21), optimal 4095M
\autoref{fig:bound_loss} (right) further shows how much smaller than the loss-optimal model is trained.
When using the loss-optimal dense Transformer as baseline, with a training budget ranging from 2.12e21 to 5.96e21 (which means the dense model has a number of parameters from 3.36B to 6.14B), an 8-expert MoE uses 23.3\% $\sim$ 28.2\% activated parameters of the loss-optimal dense model and 21.0\% $\sim$ 25.1\% of the loss-optimal 4-expert MoE. 
Two consistent trends emerge: first, as the number of experts increases, the ratio of activated parameters in the MoE model compared to the base model decreases. Second, a higher budget correlates with a lower dense model parameter ratio.
% , while other number of experts also varies in a range of 7.3\%$\sim$24.4\%. If the baseline is the loss-optimal 4-expert MoE, the ratio varies from 10.8\%$\sim$26.4\%.
% \begin{figure}[htbp]
%   \centering  \includegraphics[scale=0.35]{images/plan11_v2.pdf}
% \caption{over-train percentage under a performance bound.}
% \label{fig:over-train-ratio-bound-loss}
% \end{figure}

\paragraph{Optimal loss for bounded inference cost.} 
% Based on the scaling law, the loss $L$ is monotonic to model size $N$ before the loss-optimal point. Besides, the inference cost $I$ is also monotonic to $N$. As a result, to minimize inference cost $I$, the model size $N$ should be as low as possible, meaning the loss is as large as possible. As a result, $I_{E'}^{\min}$ corresponds to the case when the loss is exactly $L_E^{opt}$.
% Based on the above analysis, we do dichotomy search for equation $L_{E'}(N,B)=L_E^{opt}$ to find the solution $N_{E'}$, and use it to compute $I_{E'}^{\min}$. \autoref{fig:cost-under-bound-loss} shows the result for $E=1$(dense Transformer) and $4$(4-expert MoE). To reach the model performance (validation loss) similar to that of the dense model, over-training an 8-expert MoE with the same training budget results in the lowest \todo{} inference cost, which is from 38.1\% to 31.6\% as large as that of the dense model when $B$ ranges from 5.15e21 to 8.18e21.
% When the base model comes to a 4-experts MoE, both 8- and 16-experts MoE models are inference effective \todo{}, with efficiency ranging from 51.0\%  to 47.7 \% for the 8-experts MoE and from 52.0\% to 47.0\% for the 
Similarly, given a training budget $B$, we firstly compute the loss-optimal configuration for $E$-expert MoE, with its inference cost $I_E$ and $L_E$. For MoE with $E'$ experts, we compute the model size $N_{E'}$ which has an inference cost of $I_{E'}$. The monotonicity discussed before guarantees that this is the model size with the lowest loss under the inference bound. Then we use the scaling law to estimate its loss, say $L_{E'}^{\min} = Loss(N_{E'},B)$.
(the detail of the algorithm is in Algorithm \autoref{alg:bound_infer}).

% We now study what if the inference costs are bound under the baseline loss-optimal models'. For the loss-optimal $E_0$-expert setup under a training budget $B$, we compute its inference cost $C$. For every other number of experts $E$, we compute the maximal model size $N_E$ when the inference cost is no more than $C$. Then we give the optimal loss $L_E$ under training budget $B$ with a model size $N_E$.

\begin{figure}[ht]
\vspace{-1em}
  \centering   \includegraphics[scale=0.43]{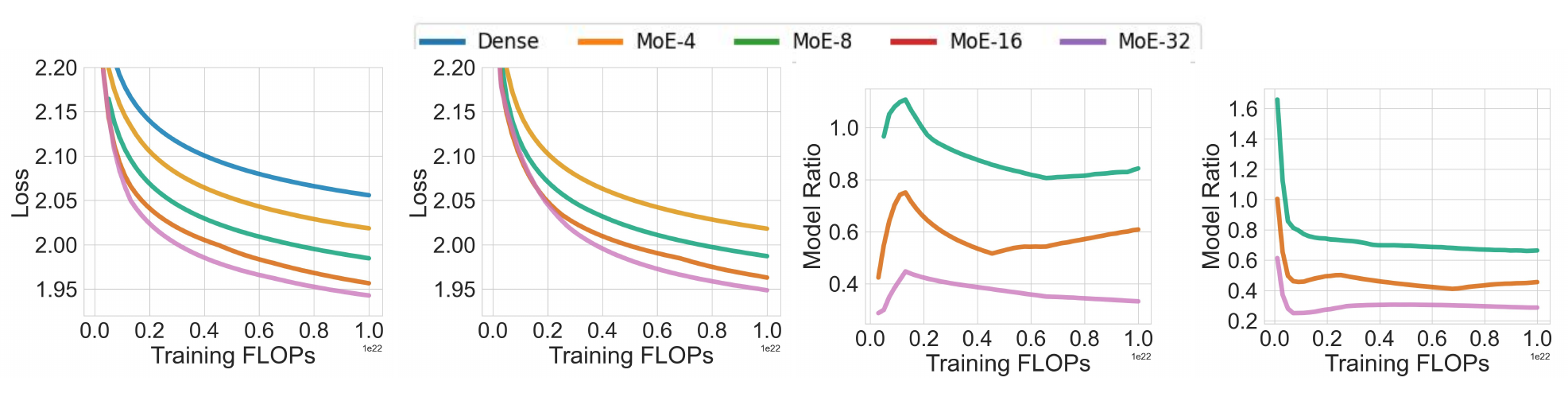}\\
    \vspace{-0.5em}
  (a) Base: Dense Model \hspace{6mm} (b) Base: MoE-4 \hspace{6mm} (c) Base: Dense Model \hspace{5mm} (d) Base: MoE-4
  \vspace{-0.5em}
\caption{\textbf{Optimal loss for a bounded inference cost.} Minimum achievable loss with a bounded inference cost (left). Ratio of model size to the base model (right).}
\label{fig:bound_infer}
\end{figure}

\autoref{fig:bound_infer} (left) shows the result when the base model is a dense Transformer or 4-experts MoE. Overtraining more experts always has a better validation loss, but the gain of scaling from 16 to 32 experts already shows a diminishing return.

Alike the bounded loss case, we also study how small is the over-trained model. \autoref{fig:bound_infer} (right) gives the ratio between the size of the over-trained $E'$-expert MoE and the loss-optimal $E$-expert MoE. When using the loss-optimal dense Transformer as the baseline, an 8-expert MoE uses 84.1\% as large as the loss-optimal base model under a training budget of 5.15e21, while other number of experts also varies in a range of 37.1\%$\sim$125.2\%. If the baseline is the loss-optimal 4-experts MoE under the same training budget, the ratio varies from 30.7\%$\sim$69.5\%, if we continue to scale the loss-optimal 4-experts MoE model, it will need 52.1\% more FLOPs in order to achieve the same loss of 8-experts MoE.

% \begin{figure}[htbp]
%   \centering \includegraphics[scale=0.35]{images/plan12_v2.pdf}
% \caption{over-train percentage under an inference cost bound.}
% \label{fig:over-train-ratio-bound-cost}
% \vspace{-1em}
% \end{figure}

\paragraph{Recommended training setup.} Over-training a smaller model with a larger dataset exhibits a great potential to reach an inference efficiency. When model quality is the most concerning factor, training a 32-experts MoE as 30\% large as the loss-optimal 32-experts MoE is preferred. If the inference cost is more important, training a 16-experts MoE as 16\% large as the loss-optimal 16-experts MoE is preferred. 
\autoref{fig:bound_loss} (right) and \autoref{fig:bound_infer} (right) prove that such a conclusion is scalable. With the growth of the training budget, the ratio of over-trained model's size against a loss-optimal model is approximately a constant.

\section{Related Work}
\vspace{-0.5em}
\paragraph{Scaling laws} Previous works extensively study the scaling behavior on different cases, especially for Transformer. \cite{kaplan2020scaling} note a power-law relationship between model size, training dataset size, and the pretrained model's quality. They suggested that when the model scales 5.5x larger, the training tokens needs to grow 1.8x larger. \cite{hoffmann2022training}, however, showed that the scaling of model size and training dataset should be scaled in equal proportions. \cite{muennighoff2023scaling} and \cite{frantar2023scaling} studied the scaling behavior for data-constrained training and sparse models, respectively, by introducing new terms to describe the data repetition and sparsity. \cite{clark2022unified} is the only attempt of MoE scaling law. It shows that MoE shows a unified scaling trend among different gating mechanisms. However, this work does not include training dataset size into consideration. As a result, unlike the later works, it cannot show a proportion between scaling model size and training dataset.
\vspace{-0.5em}
\paragraph{MoE pre-training practice} Starting from \cite{lepikhin2020gshard}, MoE architecture has been adapted with Transformer as a more cost-efficient way to scale the number of parameters. \citep{st_moe,fedus2022switch} discussed new loss function and routing mechanisms to improve the training and fine-tuning efficiency. Recent practices~\citep{dai2024deepseekmoe,jiang2024mixtral} have scaled MoE into billions of activated parameters, with a performance even stronger than the state-of-the-art Transformer models of the same size. However, these pretrained MoEs designs the hyper-parameters in an ad hoc way, simply following the scaling law of Transformers to decide the training budget allocation.
\vspace{-0.5em}
\paragraph{Budget allocation with inference cost} The closest work to this paper is \cite{sardana2023beyond}, which also recognized that inference cost should be considered in the training budget allocation problem. However, this work relied on oversimplified assumptions. It estimated inference cost with a total number of requests, which is unpredictable. Besides, it simply assumed a constant Model FLOPs Utilization (MFU) at both the training and the inference stage, while our profiling shows that MFU varies 10x with different batch sizes.

\section{Conclusion}
\vspace{-0.5em}
This paper studies the problem of how to scale the number of experts in the fast-developing MoE large language models. We first extend the scaling law, originally developed for dense transformer LLMs, to the context of MoEs, establishing a new relation between the validation loss and the number of experts, the number of training tokens, and the model size. We then discuss the need and the unique challenge to additionally consider inference efficiency when scaling MoEs. Our findings provide new insights on how to appropriately scale MoE models under compute constraints.

\clearpage
\bibliography{colm2024_conference}

\begin{thebibliography}{27}
\providecommand{\natexlab}[1]{#1}
\providecommand{\url}[1]{\texttt{#1}}
\expandafter\ifx\csname urlstyle\endcsname\relax
  \providecommand{\doi}[1]{doi: #1}\else
  \providecommand{\doi}{doi: \begingroup \urlstyle{rm}\Url}\fi

\bibitem[Brown et~al.(2020)Brown, Mann, Ryder, Subbiah, Kaplan, Dhariwal, Neelakantan, Shyam, Sastry, Askell, et~al.]{brown2020language}
Tom Brown, Benjamin Mann, Nick Ryder, Melanie Subbiah, Jared~D Kaplan, Prafulla Dhariwal, Arvind Neelakantan, Pranav Shyam, Girish Sastry, Amanda Askell, et~al.
\newblock Language models are few-shot learners.
\newblock \emph{Advances in neural information processing systems}, 33:\penalty0 1877--1901, 2020.

\bibitem[Clark et~al.(2022)Clark, de~Las~Casas, Guy, Mensch, Paganini, Hoffmann, Damoc, Hechtman, Cai, Borgeaud, van~den Driessche, Rutherford, Hennigan, Johnson, Cassirer, Jones, Buchatskaya, Budden, Sifre, Osindero, Vinyals, Ranzato, Rae, Elsen, Kavukcuoglu, and Simonyan]{clark2022unified}
Aidan Clark, Diego de~Las~Casas, Aurelia Guy, Arthur Mensch, Michela Paganini, Jordan Hoffmann, Bogdan Damoc, Blake~A. Hechtman, Trevor Cai, Sebastian Borgeaud, George van~den Driessche, Eliza Rutherford, Tom Hennigan, Matthew~J. Johnson, Albin Cassirer, Chris Jones, Elena Buchatskaya, David Budden, Laurent Sifre, Simon Osindero, Oriol Vinyals, Marc'Aurelio Ranzato, Jack~W. Rae, Erich Elsen, Koray Kavukcuoglu, and Karen Simonyan.
\newblock Unified scaling laws for routed language models.
\newblock In Kamalika Chaudhuri, Stefanie Jegelka, Le~Song, Csaba Szepesv{\'{a}}ri, Gang Niu, and Sivan Sabato (eds.), \emph{International Conference on Machine Learning, {ICML} 2022, 17-23 July 2022, Baltimore, Maryland, {USA}}, volume 162 of \emph{Proceedings of Machine Learning Research}, pp.\  4057--4086. {PMLR}, 2022.
\newblock URL \url{https://proceedings.mlr.press/v162/clark22a.html}.

\bibitem[Dai et~al.(2024)Dai, Deng, Zhao, Xu, Gao, Chen, Li, Zeng, Yu, Wu, et~al.]{dai2024deepseekmoe}
Damai Dai, Chengqi Deng, Chenggang Zhao, RX~Xu, Huazuo Gao, Deli Chen, Jiashi Li, Wangding Zeng, Xingkai Yu, Y~Wu, et~al.
\newblock Deepseekmoe: Towards ultimate expert specialization in mixture-of-experts language models.
\newblock \emph{arXiv preprint arXiv:2401.06066}, 2024.

\bibitem[Du et~al.(2022)Du, Huang, Dai, Tong, Lepikhin, Xu, Krikun, Zhou, Yu, Firat, et~al.]{du2022glam}
Nan Du, Yanping Huang, Andrew~M Dai, Simon Tong, Dmitry Lepikhin, Yuanzhong Xu, Maxim Krikun, Yanqi Zhou, Adams~Wei Yu, Orhan Firat, et~al.
\newblock Glam: Efficient scaling of language models with mixture-of-experts.
\newblock In \emph{International Conference on Machine Learning}, pp.\  5547--5569. PMLR, 2022.

\bibitem[Fedus et~al.(2022)Fedus, Zoph, and Shazeer]{fedus2022switch}
William Fedus, Barret Zoph, and Noam Shazeer.
\newblock Switch transformers: Scaling to trillion parameter models with simple and efficient sparsity.
\newblock \emph{The Journal of Machine Learning Research}, 23\penalty0 (1):\penalty0 5232--5270, 2022.

\bibitem[Frantar et~al.(2023)Frantar, Riquelme, Houlsby, Alistarh, and Evci]{frantar2023scaling}
Elias Frantar, Carlos Riquelme, Neil Houlsby, Dan Alistarh, and Utku Evci.
\newblock Scaling laws for sparsely-connected foundation models.
\newblock \emph{arXiv preprint arXiv:2309.08520}, 2023.

\bibitem[Hoffmann et~al.(2022)Hoffmann, Borgeaud, Mensch, Buchatskaya, Cai, Rutherford, Casas, Hendricks, Welbl, Clark, et~al.]{hoffmann2022training}
Jordan Hoffmann, Sebastian Borgeaud, Arthur Mensch, Elena Buchatskaya, Trevor Cai, Eliza Rutherford, Diego de~Las Casas, Lisa~Anne Hendricks, Johannes Welbl, Aidan Clark, et~al.
\newblock Training compute-optimal large language models.
\newblock \emph{arXiv preprint arXiv:2203.15556}, 2022.

\bibitem[Huber(1992)]{huber1992robust}
Peter~J Huber.
\newblock Robust estimation of a location parameter.
\newblock In \emph{Breakthroughs in statistics: Methodology and distribution}, pp.\  492--518. Springer, 1992.

\bibitem[Jacobs et~al.(1991)Jacobs, Jordan, Nowlan, and Hinton]{jacobs1991adaptive}
Robert~A Jacobs, Michael~I Jordan, Steven~J Nowlan, and Geoffrey~E Hinton.
\newblock Adaptive mixtures of local experts.
\newblock \emph{Neural computation}, 3\penalty0 (1):\penalty0 79--87, 1991.

\bibitem[Jiang et~al.(2024)Jiang, Sablayrolles, Roux, Mensch, Savary, Bamford, Chaplot, Casas, Hanna, Bressand, et~al.]{jiang2024mixtral}
Albert~Q Jiang, Alexandre Sablayrolles, Antoine Roux, Arthur Mensch, Blanche Savary, Chris Bamford, Devendra~Singh Chaplot, Diego de~las Casas, Emma~Bou Hanna, Florian Bressand, et~al.
\newblock Mixtral of experts.
\newblock \emph{arXiv preprint arXiv:2401.04088}, 2024.

\bibitem[Jordan \& Jacobs(1994)Jordan and Jacobs]{jordan1994hierarchical}
Michael~I Jordan and Robert~A Jacobs.
\newblock Hierarchical mixtures of experts and the em algorithm.
\newblock \emph{Neural computation}, 6\penalty0 (2):\penalty0 181--214, 1994.

\bibitem[Kaplan et~al.(2020)Kaplan, McCandlish, Henighan, Brown, Chess, Child, Gray, Radford, Wu, and Amodei]{kaplan2020scaling}
Jared Kaplan, Sam McCandlish, Tom Henighan, Tom~B Brown, Benjamin Chess, Rewon Child, Scott Gray, Alec Radford, Jeffrey Wu, and Dario Amodei.
\newblock Scaling laws for neural language models.
\newblock \emph{arXiv preprint arXiv:2001.08361}, 2020.

\bibitem[Kwon et~al.(2023)Kwon, Li, Zhuang, Sheng, Zheng, Yu, Gonzalez, Zhang, and Stoica]{kwon2023efficient}
Woosuk Kwon, Zhuohan Li, Siyuan Zhuang, Ying Sheng, Lianmin Zheng, Cody~Hao Yu, Joseph Gonzalez, Hao Zhang, and Ion Stoica.
\newblock Efficient memory management for large language model serving with pagedattention.
\newblock In \emph{Proceedings of the 29th Symposium on Operating Systems Principles}, pp.\  611--626, 2023.

\bibitem[Lepikhin et~al.(2020)Lepikhin, Lee, Xu, Chen, Firat, Huang, Krikun, Shazeer, and Chen]{lepikhin2020gshard}
Dmitry Lepikhin, HyoukJoong Lee, Yuanzhong Xu, Dehao Chen, Orhan Firat, Yanping Huang, Maxim Krikun, Noam Shazeer, and Zhifeng Chen.
\newblock Gshard: Scaling giant models with conditional computation and automatic sharding.
\newblock \emph{arXiv preprint arXiv:2006.16668}, 2020.

\bibitem[Liu \& Nocedal(1989)Liu and Nocedal]{liu1989limited}
Dong~C Liu and Jorge Nocedal.
\newblock On the limited memory bfgs method for large scale optimization.
\newblock \emph{Mathematical programming}, 45\penalty0 (1-3):\penalty0 503--528, 1989.

\bibitem[Loshchilov \& Hutter(2017)Loshchilov and Hutter]{loshchilov2017decoupled}
Ilya Loshchilov and Frank Hutter.
\newblock Decoupled weight decay regularization.
\newblock \emph{arXiv preprint arXiv:1711.05101}, 2017.

\bibitem[Muennighoff et~al.(2023)Muennighoff, Rush, Barak, Scao, Piktus, Tazi, Pyysalo, Wolf, and Raffel]{muennighoff2023scaling}
Niklas Muennighoff, Alexander~M Rush, Boaz Barak, Teven~Le Scao, Aleksandra Piktus, Nouamane Tazi, Sampo Pyysalo, Thomas Wolf, and Colin Raffel.
\newblock Scaling data-constrained language models.
\newblock \emph{arXiv preprint arXiv:2305.16264}, 2023.

\bibitem[Narayanan et~al.(2021)Narayanan, Shoeybi, Casper, LeGresley, Patwary, Korthikanti, Vainbrand, Kashinkunti, Bernauer, Catanzaro, et~al.]{narayanan2021efficient}
Deepak Narayanan, Mohammad Shoeybi, Jared Casper, Patrick LeGresley, Mostofa Patwary, Vijay Korthikanti, Dmitri Vainbrand, Prethvi Kashinkunti, Julie Bernauer, Bryan Catanzaro, et~al.
\newblock Efficient large-scale language model training on gpu clusters using megatron-lm.
\newblock In \emph{Proceedings of the International Conference for High Performance Computing, Networking, Storage and Analysis}, pp.\  1--15, 2021.

\bibitem[Narayanan et~al.(2023)Narayanan, Santhanam, Henderson, Bommasani, Lee, and Liang]{narayanan2023cheaply}
Deepak Narayanan, Keshav Santhanam, Peter Henderson, Rishi Bommasani, Tony Lee, and Percy Liang.
\newblock Cheaply estimating inference efficiency metrics for autoregressive transformer models.
\newblock In \emph{Thirty-seventh Conference on Neural Information Processing Systems}, 2023.

\bibitem[Rasley et~al.(2020)Rasley, Rajbhandari, Ruwase, and He]{rasley2020deepspeed}
Jeff Rasley, Samyam Rajbhandari, Olatunji Ruwase, and Yuxiong He.
\newblock Deepspeed: System optimizations enable training deep learning models with over 100 billion parameters.
\newblock In \emph{Proceedings of the 26th ACM SIGKDD International Conference on Knowledge Discovery \& Data Mining}, pp.\  3505--3506, 2020.

\bibitem[Sardana \& Frankle(2023)Sardana and Frankle]{sardana2023beyond}
Nikhil Sardana and Jonathan Frankle.
\newblock Beyond chinchilla-optimal: Accounting for inference in language model scaling laws.
\newblock \emph{arXiv preprint arXiv:2401.00448}, 2023.

\bibitem[Shazeer et~al.(2017)Shazeer, Mirhoseini, Maziarz, Davis, Le, Hinton, and Dean]{shazeer2017outrageously}
Noam Shazeer, Azalia Mirhoseini, Krzysztof Maziarz, Andy Davis, Quoc Le, Geoffrey Hinton, and Jeff Dean.
\newblock Outrageously large neural networks: The sparsely-gated mixture-of-experts layer.
\newblock \emph{arXiv preprint arXiv:1701.06538}, 2017.

\bibitem[Shoeybi et~al.(2019)Shoeybi, Patwary, Puri, LeGresley, Casper, and Catanzaro]{shoeybi2019megatron}
Mohammad Shoeybi, Mostofa Patwary, Raul Puri, Patrick LeGresley, Jared Casper, and Bryan Catanzaro.
\newblock Megatron-lm: Training multi-billion parameter language models using model parallelism.
\newblock \emph{arXiv preprint arXiv:1909.08053}, 2019.

\bibitem[Smith et~al.(2022)Smith, Patwary, Norick, LeGresley, Rajbhandari, Casper, Liu, Prabhumoye, Zerveas, Korthikanti, et~al.]{smith2022using}
Shaden Smith, Mostofa Patwary, Brandon Norick, Patrick LeGresley, Samyam Rajbhandari, Jared Casper, Zhun Liu, Shrimai Prabhumoye, George Zerveas, Vijay Korthikanti, et~al.
\newblock Using deepspeed and megatron to train megatron-turing nlg 530b, a large-scale generative language model.
\newblock \emph{arXiv preprint arXiv:2201.11990}, 2022.

\bibitem[Soboleva et~al.(2023)Soboleva, Al-Khateeb, Myers, Steeves, Hestness, and Dey]{cerebras2023slimpajama}
Daria Soboleva, Faisal Al-Khateeb, Robert Myers, Jacob~R Steeves, Joel Hestness, and Nolan Dey.
\newblock {SlimPajama: A 627B token cleaned and deduplicated version of RedPajama}.
\newblock \url{https://www.cerebras.net/blog/slimpajama-a-627b-token-cleaned-and-deduplicated-version-of-redpajama}, June 2023.
\newblock URL \url{https://huggingface.co/datasets/cerebras/SlimPajama-627B}.

\bibitem[Touvron et~al.(2023)Touvron, Lavril, Izacard, Martinet, Lachaux, Lacroix, Rozi{\`e}re, Goyal, Hambro, Azhar, et~al.]{touvron2023llama}
Hugo Touvron, Thibaut Lavril, Gautier Izacard, Xavier Martinet, Marie-Anne Lachaux, Timoth{\'e}e Lacroix, Baptiste Rozi{\`e}re, Naman Goyal, Eric Hambro, Faisal Azhar, et~al.
\newblock Llama: Open and efficient foundation language models.
\newblock \emph{arXiv preprint arXiv:2302.13971}, 2023.

\bibitem[Zoph(2022)]{st_moe}
Barret Zoph.
\newblock Designing effective sparse expert models.
\newblock In \emph{{IEEE} International Parallel and Distributed Processing Symposium, {IPDPS} Workshops 2022, Lyon, France, May 30 - June 3, 2022}, pp.\  1044. {IEEE}, 2022.
\newblock URL \url{https://doi.org/10.1109/IPDPSW55747.2022.00171}.

\end{thebibliography}
\bibliographystyle{colm2024_conference}

\appendix
\section{Optimal Allocation For Dense Model}
\label{appendix:opt_alloc}
Given that the training budget can be approximated as $C=6ND$ ~\citep{kaplan2020scaling} and the optimal allocation problem illustrated in  \autoref{eq:optimal_allocation}, we can solve this convex optimization problem with an equality constraint by adding a Lagrange multiplier.

\begin{equation*}
    \begin{aligned} 
    \mathcal{L}(N, D, \lambda) = L(N,D) = L_0 + \frac{A}{N^\alpha} + \frac{B}{D^\beta} + \lambda (6ND - C)
    \end{aligned}
    \label{eq:lagrange_opt_alloc}
\end{equation*}
The dual problem is $g(\lambda) = \inf_{N, D}\mathcal{L}(N, D, \lambda)$.

By taking the derivative with respect to $N$ and $D$, we have:

\begin{equation*}
\begin{aligned}
\frac{\partial \mathcal{L}}{\partial N} &= \frac{\partial}{\partial N} \left( L_0 + \frac{A}{N^\alpha} + \frac{B}{D^\beta} + \lambda (6ND - C) \right) \\
&= \frac{\partial}{\partial N} \left( \frac{A}{N^\alpha} \right) + \frac{\partial}{\partial N} \left( \lambda (6ND - C) \right) \\
&= -\frac{\alpha A}{N^{\alpha+1}} + 6\lambda D
\end{aligned}
\end{equation*}

\begin{equation*}
\begin{aligned}
\frac{\partial \mathcal{L}}{\partial D} &= \frac{\partial}{\partial D} \left( L_0 + \frac{A}{N^\alpha} + \frac{B}{D^\beta} + \lambda (6ND - C) \right) \\
&= \frac{\partial}{\partial D} \left( \frac{B}{D^\beta} \right) + \frac{\partial}{\partial D} \left( \lambda (6ND - C) \right) \\
&= -\frac{\beta B}{D^{\beta+1}} + 6\lambda N
\end{aligned}
\end{equation*}

Let both the derivatives equal 0 and also apply the constraint $C=6ND$. We can calculate the loss-optimal configuration as shown in  \autoref{eq:openai_scaling_law_opt}.
\section{Training Details}
\vspace{-2em}
\label{appendix:training_details}
\paragraph{Model Details} 
\begin{figure}[H]
  \centering   \includegraphics[scale=0.35]{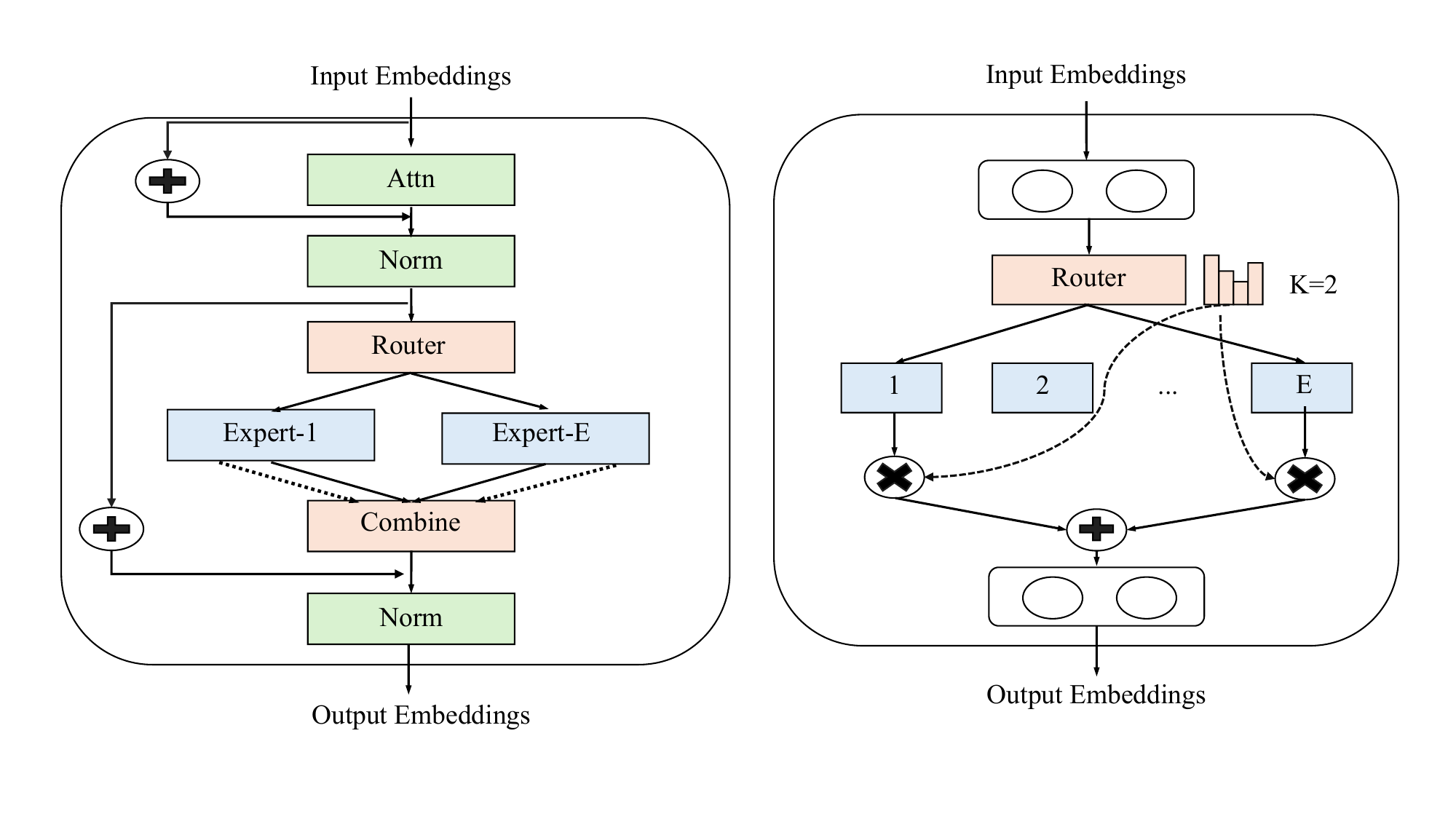}
  \vspace{-2em}
\caption{MoE architecture.}
\label{fig:moe-arch}
\end{figure}
 As seen in \autoref{fig:moe-arch}, a Transformer's MoE layer is composed of $E$ feed-forward networks, labeled $\mathrm{FFN}_1$ to $\mathrm{FFN}_E$. Given an input token $u_t^l$ (i.e. logits of token $t$ in the $l$-th layer) to this MoE layer, its output is a sum of the outputs from these experts, calculated as $\sum_{e=1}^E \mathcal{G}{i, t} \cdot \operatorname{FFN}e\left(u_t^l\right)$. Here, $\mathcal{G}{i, t}$ is a vector determined by a gating mechanism $\operatorname{GATE}(\cdot)$. It's decided that each token is routed to no more than $K$ experts, which causes the gating values $\mathcal{G}_{i, t}$ to be non-zero for those experts involved, indicating their respective contributions to the overall output of the network.

\begin{equation*}
    \begin{aligned} \mathbf{h}_t^l & =\sum_{i=1}^E\left(\mathcal{G}_{i, t} \operatorname{FFN}_i\left({u}_t^l\right)\right)+{u}_t^l, \\ \mathcal{G}_{i, t}& = \begin{cases}s_{i, t}, & s_{i, t} \in \operatorname{Topk}\left(\left\{s_{j, t} \mid 1 \leqslant j \leqslant E\right\}, K\right), \\ 0, & \text { otherwise }\end{cases} \\ s_{i, t} & =\operatorname{Softmax}_i\left({u}_t^{{l}^T}\right),\end{aligned}
\end{equation*}
% here $u_t^l$ is the hidden state of the token after the $l$-th attention module, $h_t^l$ represents the output, $g_{i,t}$ denotes the gate value for the $i$-th expert, $s_{i, t}$ represents the affinity between token and the $i$-th expert, and $e_i^l$ is the centroid of the $i$-th expert in the $l$-th layer. It's important to note that Topk refers to the group that contains the $K$ highest affinity scores. These scores are calculated for the 
% $t$-th token against all $E$ experts.

Recent works~\citep{du2022glam, st_moe, fedus2022switch, lepikhin2020gshard} suggest to replace one of every two FFN layers in a Transformer model by MoE, and use Top-2 gating~\citep{shazeer2017outrageously} as the routing mechanism. In this paper, we also inherit from such a context. Besides, our model architecture follows the practice of Llama~\citep{touvron2023llama}, which uses a gated-MLP as the feed-forward layer, and the MLP intermediate hidden dimension size is 2.6x large as the model's hidden dimension.

To train our model, we have forked Megatron-Deepspeed ~\citep{rasley2020deepspeed, smith2022using} framework. Models are trained using data, tensor parallelism on up to 32 GPUs.

\paragraph{Dataset} 
we specify our dataset choice as SlimPajama ~\citep{cerebras2023slimpajama}, a high-quality dataset refined through content filtering and deduplication processes. 
It is an open-source version of the LLaMA pretraining data blend, comprising 82\% internet content (with 67\% from CommonCrawl and 15\% from C4), 4.5\% code (sourced from Github), 4.5\% from Wikipedia, 4.5\% from books, 2.5\% from Arxiv, and 2\% from StackExchange.
Given that this dataset closely resembles the one used for pretraining LLaMA models, there is less concern about adapting the findings to various datasets.
From this dataset, our experiments utilize up to 20 billion tokens for training and 0.58 billion tokens for validation purposes.

\paragraph{Training Details}
\label{sec:training-details}
All models were trained on A100 GPUs, utilizing a blend of data, tensor, and model parallelism as outlined in \cite{shoeybi2019megatron, narayanan2021efficient}. The training involved a sequence length of 2048 and a batch size of 256 (i.e. 0.5M tokens per batch). All models are optimized with AdamW~\citep{loshchilov2017decoupled}. Due to empirical observations, it has been determined that larger models necessitate a reduced learning rate to avoid divergence, whereas smaller models can withstand a higher learning rate. Consequently, we establish the learning rate based on previous experience~\citep{kaplan2020scaling}:
\begin{equation*}
    \operatorname{LR}(N) \approx 0.003239+-0.0001395 \log (N)
\end{equation*}
We also employ a linear warm-up of the learning rate with the initial 3\% tokens. The learning rate then decays to 10\% of the maximum value through a cosine schedule.
% More details of the training are mentioned in \autoref{sec:appendix-scaling-experiment-details}.
\section{Inference Cost Estimation}
\label{appendix:inference_cost}
\paragraph{Model size}
The model size of MoE model refers to the size of the corresponding dense model, as described in Section~\ref{sec:mixture-of-expert-model}. The total number of parameters can be approximately described as proportional to $N * (1 + (E - 1) c)$, where the factor $c$ is influenced by the model architecture. In our setup, we replace a FFN layer by MoE for every two Transformer layers. The width of the Gated-MLP layers $i$ is fixed at around 2.67 times the width of the model hidden state $h$~\citep{touvron2023llama}, so FFN layers take $2/3$ of all parameters in the dense model. Consequently, $c$ equals $1/3$.

In a Transformer layer, the parameter count primarily stems from two components: the self-attention module and the feed-forward network.

Within the self-attention mechanism, four matrices of parameters exist: $W_k, W_v, W_q, W_o$, each having dimensions $h\times h$. Additionally, the bias components contribute $4h$ parameters. Therefore, the self-attention mechanism altogether encompasses $4h^2 + 4h$ parameters.

Regarding the gated MLP, there are three linear projections involved: the gate projection, which is $h \times 2.67h$; the up projection, also $h \times 2.67h$; and the down projection, which is $2.67h \times h$. Consequently, the MLP component holds a total of $8.01h^2 + 6.34h$ parameters.

Excluding the linear term, the proportion of parameters attributed to the MLP relative to the total is approximately $\frac{8.01}{8.01 + 6.34} \approx \frac{2}{3}$.

\paragraph{Maximal batch size} For every token processed, the KV-cache memory for a token is $2hl$, with the hidden dimension size $h$ and the number of layers $l$. Assume a model has $N_m$ parameters, each GPU has $M_0$ memory, the available memory for KV-cache is $GM_0-N_m$. 
Assume that the average output length is $n$, and the average prompt length is $p$. The memory for a single request's KV-cache grows from $2phl$ to $2(n+p)hl$, and the expectation is $(2p+n)hl$. Hence, the maximum number of simultaneous requests that can be served is given by $b=\frac{GM_0-N_m}{(2p+n)hl}$.

\paragraph{Latency} When serving with a batch size $b$, the decoding iteration's batch size is $b$. Given the average output length $n$, we can expect that on average, $b/n$ requests will be completed in a decoding iteration. On the other hand, to maintain the batch size stable, it needs $b/n$ new requests, necessitating an additional prompt iteration. Hence, the latency per iteration for model $m$ has:
\begin{equation*}
    L_m(b,G) = L_{m}^P(b/n,G) + L_{m}^D(b,G)
\end{equation*}

were $L_{m}^P(b,G),L_{m}^D(b,G)$ are the prompt and decoding latency with a batch size $b$ on $G$ GPUs. The prompt and decoding stages exhibit distinct levels of computing intensities. To assess the latency of each stage, we separately conduct a detailed profiling of various models for each stage. This data is used to estimate latency for other models through linear interpolation on batch size and model size.

\paragraph{Throughput} Let $k=2p+n$, the throughput $T_m$ of model $m$ has:
\begin{equation}
    T_m =\frac{GM_0-N_m}{khl(L_{m}^P(\frac{GM_0-N_m}{knhl},G) + L_{m}^D(\frac{GM_0-N_m}{khl},G))} 
    \label{eq:throughput}
\end{equation}

Since $p$ and $n$ depend solely on the request's traffic patterns, together with $k$ is a constant. We approximate their values with the ShareGPT dataset. 

\begin{figure}[h!] 
  \centering \includegraphics[scale=0.30]{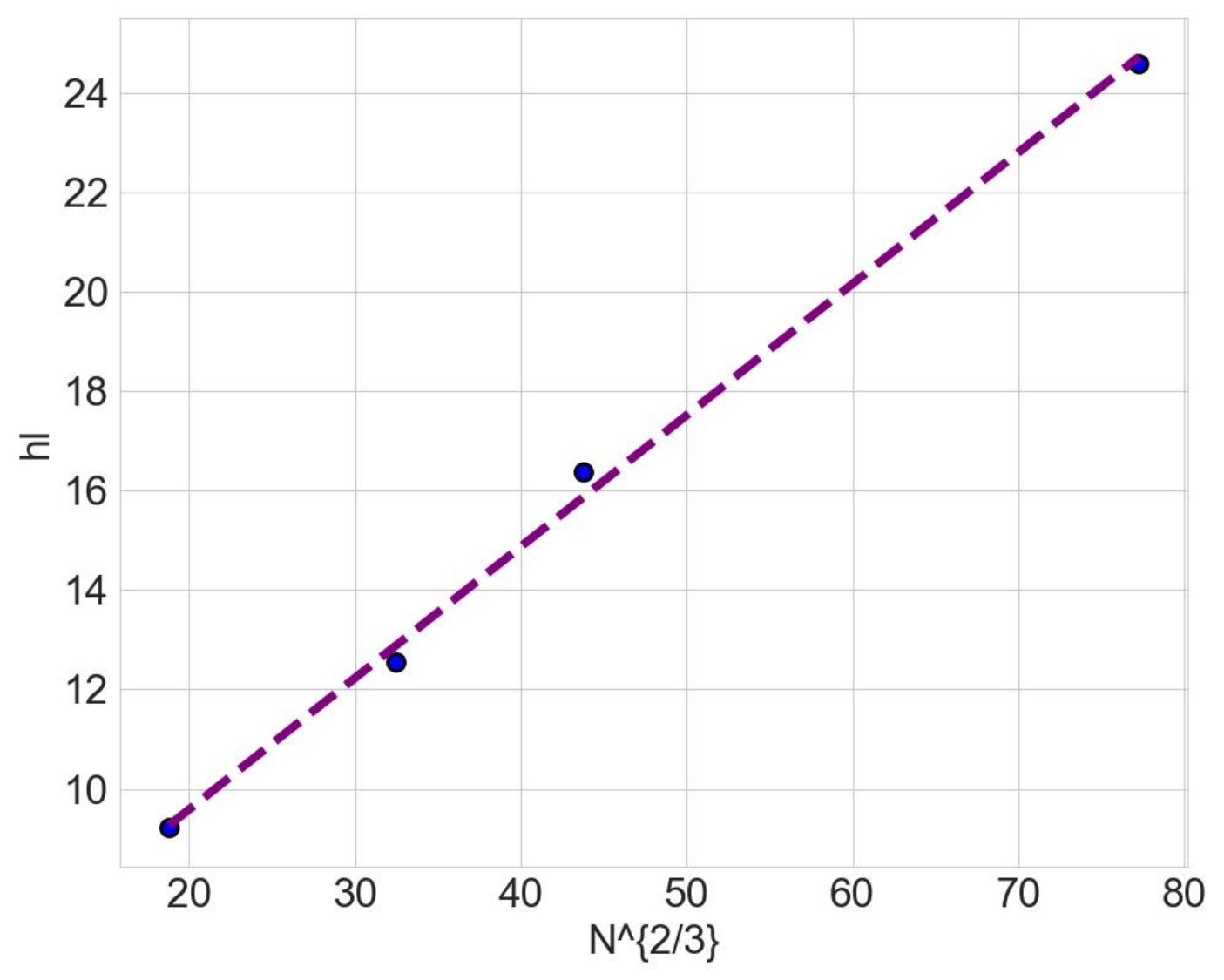}
  \vspace{-1em}
  \caption{\textbf{Fitted hl with $N^{2/3}$.} Dots represent the actual $hl$ value, the line indicate the fitted value.}
  \vspace{-2em}
  \label{fig:fit_hm_lm} 
\end{figure}

Furthermore, there is $N\propto h^2l$. To estimate the $hl$ term in the model's throughput, we take a simple assumption that hidden state and number of layers roughly keep a linear relationship. As a result, there is $hl=\mu N^{2/3}$, where $\mu$ is a constant. As \autoref{fig:fit_hm_lm} shows, the accurate predicted $hl$ assures that our assumption is reasonable.

\section{Detail of Fitting the Scaling Law}
\label{sec:appendix-scaling-experiment-details}

% Except for the small token dispatching overhead, MoE execution cost is $(1 + (K-1)c)$ times as the dense model. Since we use Top-2 gating in MoE layers, here $K$ equals $2$.

To estimate $(\alpha, \beta, \gamma, A, B, C, d, F)$, we effectively minimize the Huber loss~\citep{huber1992robust}:
\begin{align*}
&\min_{A, B, C, d, F, \alpha, \beta, \gamma} \sum_{\text{Run } i} \text{Huber}_\delta\left(\log \hat{L}\left(N_i, D_i, \hat{E_i}\right)-\log L_i\right)
\end{align*}
% \\ &= \min_{A, B, C, d, F, \alpha, \beta, \gamma} \sum_{\text{Run } i} \text{Huber}_\delta (\text{LSE}(a - \alpha \log N, b - \beta \log E, \\
% &\quad c - \gamma \log D + d \log N \log E, f))
% where \emph{LSE} is the log-sum-exp operator, and we set $A,B,C,F=exp(a),exp(b),exp(c),exp(f)$.

We use the L-BFGS~\citep{liu1989limited} algorithm to find local minima of the objective above, started on a grid of initialisation given by: $\alpha \in \{0., 0.5, ..., 2.\}, \beta \in \{0., 0.5, ..., 2.\}, \gamma \in \{0., 0.5, ..., 2.\}, a \in \{0, 5, ..., 25\}, b \in \{0, 5, ..., 25\},  c \in \{0, 5, ..., 25\}, d \in \{0, 5, ..., 25\}, f \in \{\-1., -.5, ..., 1.\}$.
We use $\delta=10^{-3}$ for the Huber loss, which is robust shown in previous work~\citep{hoffmann2022training}.

We also compute RMSLE value and Huber loss value, which are 3.908e-3 and 1.033e-3, respectively, indicating that the error is extremely low.

% \begin{table}[ht!]
% \centering
% \caption{Parameters of Dense-equivalent Sclaing Law for MoE with a fixed number of experts}
% \begin{tabular}{c | c c c c c}
%  \toprule
%   MoE model 
%   &
%   $\alpha$
%   & $\beta$
%   & $A$
%   & $B$
%   & $E$
%   \\ [0.5ex] 
%   \midrule
%   % alpha: 0.35919639611542237 beta: 0.44692951129854924 A: 349.98828125, B: 11692.892578125, F: 1.7920018434524536
%   Ep-8 & 0.359 & 0.447 & 349.988 & 11692.893 & 1.792\\ 
%   \midrule
%   % alpha: 0.38748073955678014 beta: 0.4287698272051654 A: 520.3478393554688, B: 8223.376953125, F: 1.7799242734909058
%   Ep-16 & 0.387 & 0.429 & 520.348 & 8223.377 & 1.780\\ 
%   \bottomrule
% \end{tabular}
% \end{table}

\section{Bound Metrics} 
Here we provide the detail algorithm for \autoref{sec:scaling-behavior} about studying the optimal inference cost under a bounded loss, or optimal loss under an inference cost. In both cases, the bound is defined by the loss-optimal MoE with fewer experts.
% More specifically, given a training budget $B$, we first find the loss-optimal budget allocation $(N_E,D_E)$ under a fixed number of experts $E$. The validation loss and inference cost for this model is correspondingly $L_E^{opt}$ and $I_E$. Then for MoE with a larger number of expert $E'>E$, we study its over-trained configuration, where its quality is anchored by $L_E$, say $L_{E'}\le L_E^{opt}$. We compute the lowest inference cost $I_{E'}^{\min}$ for $E'$ experts under the quality constraint above, and compare $I_{E'}$ with $I_E$. On the other hand, we also consider the lowest validation loss $L_{E'}^{\min}$, under a bound that $I_{E'}\le I_E$, and compare $L_{E'}^{\min}$ with $L_E^{opt}$.

% \begin{wrapfigure}{b}{0.48\textwidth}
% \begin{minipage}{0.48\textwidth}
\begin{algorithm}[H]
\caption{Optimal Inference Cost For A Bounded Loss.}
\label{alg:bound_loss}
\textbf{Input}: A training budget $B$ \\
\hspace*{3\algorithmicindent}   A base model with number of experts $E$\\
\hspace*{3\algorithmicindent}   A MoE model with a larger number of experts $E'$\\
\hspace*{3\algorithmicindent}   Total GPU number $G$ \\
%\hspace*{2\algorithmicindent}  steps \\
\textbf{Output}: Lowest inference cost $I_{E'}^{\min}$  \\
\begin{algorithmic}[1]
 \STATE $(N_E, D_E) \gets \texttt{Optimal\_config}(B)$
 \STATE $L_E \gets \texttt{Scaling\_law}(N_E, D_E, E)$
 \FOR{$g \gets 1$ \textbf{to} $G$}
    \STATE $I_E \gets \min (\texttt{Get\_cost}(N_E, E, g), I_E)$
\ENDFOR
\STATE $N_{E'} \gets \texttt{Dichotomy\_search}(E', L_E)$
\FOR{$g \gets 1$ \textbf{to} $G$}
    \STATE $I_{E'}^{\min} \gets \min (\texttt{Get\_cost}(N_{E'}, E', g), I_{E'}^{\min})$
\ENDFOR
  \RETURN{$I_{E'}^{\min}$}
\end{algorithmic}
\end{algorithm}

\begin{algorithm}[H]
\caption{Optimal Loss For A Bounded Inference Cost.}
\label{alg:bound_infer}
\textbf{Input}: A training budget $B$ \\
\hspace*{3\algorithmicindent}   A base model with number of experts $E$\\
\hspace*{3\algorithmicindent}   A MoE model with a larger number of experts $E'$\\
\hspace*{3\algorithmicindent}   Total GPU number $G$ \\
%\hspace*{2\algorithmicindent}  steps \\
\textbf{Output}: Lowest validation loss $L_{E'}^{\min}$  \\
\begin{algorithmic}[1]
 \STATE $(N_E, D_E) \gets \texttt{Optimal\_config}(B)$ 
 % \STATE $L_E \gets \texttt{Scaling\_law}(N_E, D_E, E)$
 \FOR{$g \gets 1$ \textbf{to} $G$}
    \STATE $I_E \gets \min (\texttt{Get\_cost}(N_E, E, g), I_E)$
\ENDFOR
\STATE $N_{E'} \gets \texttt{Dichotomy\_search}(E', I_E)$
\STATE $D_{E'} \gets \texttt{Dataset\_size} (B, N_{E'})$
 \STATE $L_{E'} \gets \texttt{Scaling\_law}(N_{E'}, D_{E'}, E')$
% \FOR{$g \gets 1$ \textbf{to} $8$}
%     \STATE $I_{E'}^{\min} \gets \min (\texttt{Get\_cost}(N_{E'}, E', g), I_{E'}^{\min})$
% \ENDFOR
  \RETURN{$L_{E'}^{\min}$}
\end{algorithmic}
\end{algorithm}
% \end{minipage}
% \end{wrapfigure}
% \begin{wrapfigure}{b}{0.48\textwidth}
% \begin{minipage}{0.48\textwidth}
% \begin{algorithm}[H]
% \caption{Projected Gradient Descent}
% \label{alg1}
% \textbf{Input}: A labeled example $(x, y)$ \\
% \hspace*{2\algorithmicindent}    a perturbation amount $\epsilon$ and total steps \\
% %\hspace*{2\algorithmicindent}  steps \\
% \textbf{Output}: An adversarial example $\Tilde{x}$ \\
% \begin{algorithmic}[1]
% \small
%  \STATE $\alpha \gets \epsilon / \texttt{steps}$ 
%  \STATE $ori\_x \gets \texttt{x}$
% \FOR{$i \gets 1$ \textbf{to} $\texttt{steps}$}
%     \STATE $\texttt{grad} \gets \texttt{MLP.gradient}(x, y)$ 
%     \STATE $x \gets x + \alpha * \texttt{sign}(\texttt{grad})$
%     \STATE $\eta \gets \texttt{clip}(x - ori\_x, -\epsilon, \epsilon)$
%     \STATE $x \gets \texttt{clip}(ori\_x + \epsilon, 0, 1)$
% \ENDFOR
%   % \WHILE{$|\mathcal{D}| > 0$}
%   %   \STATE $d_i \gets \texttt{min\_deg}(\mathcal{D})$  
%   %   % \footnotesize\ttfamily\textcolor{blue}{\# Return a min-degree unvisited doc}
%   %   \STATE $p \gets [d_i]$  \COMMENT{Initialize path $p$ from $d_i$}
%   %   \WHILE{$\exists d_j \in N(d_i)$ in $\mathcal{D}$ }
%   %       \STATE $d_i \gets d_j$
%   %       \STATE $p.append(d_i)$
%   %       \STATE $\mathcal{D}.remove(d_i)$
%   %    \ENDWHILE
%   %   \STATE $P.append(p)$
%   % \ENDWHILE
%   % \RETURN{List of paths $P=\{p\}$ covering the graph}
%   \RETURN{$x$}
% \end{algorithmic}
% \end{algorithm}
% \end{minipage}
% \end{wrapfigure}
\end{document}